\documentclass[letterpaper, 10pt, conference]{ieeeconf}

\IEEEoverridecommandlockouts
\usepackage{cite}
\usepackage{amsmath,amssymb,amsfonts}
\usepackage{graphicx}
\usepackage{textcomp}
\usepackage{xcolor}

\usepackage{graphics} 
\usepackage{epsfig} 
\usepackage{mathptmx} 
\usepackage{times} 
\usepackage{amsmath} 
\usepackage{amssymb}  
\usepackage{subfigure}
\usepackage{multirow}
\usepackage{subcaption}
\usepackage{float}
\usepackage{ctable}
\usepackage{cuted}
\usepackage{colortbl}
\usepackage{algorithm}
\usepackage{algpseudocode}
\usepackage{titlesec}
\usepackage{soul}
\usepackage{pifont}
\definecolor{myblue}{HTML}{E8F2F6}

\usepackage[numbers,sort&compress]{natbib}
\usepackage{eucal} 
\usepackage[colorlinks,linkcolor=blue]{hyperref}

\usepackage{etoolbox}

\AtBeginDocument{%
    \apptocmd{\thebibliography}{}{}
}

\def\BibTeX{{\rm B\kern-.05em{\sc i\kern-.025em b}\kern-.08em
    T\kern-.1667em\lower.7ex\hbox{E}\kern-.125emX}}
\begin{document}

\title{{Touch2Robot: Robot Touch in the Human Demonstration Loop}}

\author{\renewcommand{\arraystretch}{1.35}\begin{tabular}{c}
    Shengcheng Luo\textsuperscript{1,2*}, Xiaoyang Chen\textsuperscript{1,2,3*}, Hong Ying\textsuperscript{1*}, Xiaoying Zhou\textsuperscript{1*},\\
    Jiaming Jiang\textsuperscript{1}, Haoran Guo\textsuperscript{1}, Wanlin Li\textsuperscript{2}, Ziyuan Jiao\textsuperscript{2,4$\dagger$}, Chenxi Xiao\textsuperscript{1$\dagger$}
  \end{tabular}
\thanks{* Equal contribution. $\dagger$ Corresponding author.}
\thanks{$^{1}$ShanghaiTech University, $^{2}$Beijing Institute for General Artificial Intelligence (BIGAI), $^{3}$Shanghai Jiao Tong University, $^{4}$Beihang University. }
}



\maketitle


\begin{abstract}
Human demonstrations offer a scalable way to collect manipulation data, but their contacts may be unstable or infeasible when transferred to a robot hand. Collecting demonstrations directly on the target robot avoids this mismatch, but substantially increases the cost of data collection.
To address this trade-off, we present \textbf{Touch2Robot}, a framework that lets humans collect demonstrations while seeing how the target robot hand would contact the object. We capture human hand motion, tactile-glove measurements, and object motion during human manipulation. These recordings guide object-specific RL policies to reproduce the demonstrated object motion while favoring contacts consistent with the recorded human touch. We distill the learned behaviors into a unified real-time retargeter that maps incoming human observations and object geometry to robot hand configurations. During collection, the predicted robot configuration is synchronized with the tracked object pose in simulation to reconstruct robot-object contacts, which are visualized to help the demonstrator adapt subsequent interactions to the target hand.
Across four real-world tasks, Touch2Robot improves average real-robot replay completion from 37.9\% to 72.1\% over visual-only feedback, while reducing the collection time per replay-successful demonstration from 58.6~s to 18.2~s. Reconstructed target-hand contacts achieve 44.2\% F1 against real-robot tactile measurements, and policies trained on Touch2Robot demonstrations improve downstream Diffusion Policy performance by 29.1 percentage points over visual-only feedback. These results show that bringing robot touch into the human demonstration loop improves both the quality and efficiency of scalable dexterous data collection.
\textit{Project webpage: \href{https://Touch2Robot.github.io/}{https://Touch2Robot.github.io/}.}
\end{abstract}

\section{INTRODUCTION}
Dexterous manipulation relies on demonstrations that establish, maintain, and release object contacts compatible with the target robot's embodiment \cite{dexmani,blind,zhu2026learningdexterousmanipulationusing}. Human demonstrations provide an inexpensive and scalable source of manipulation data, but their contact patterns are shaped by the morphology and kinematics of the human hand. Consequently, contacts that are stable and effective for a human can be unstable or infeasible for the target robot. Although collecting demonstrations directly on the robot captures embodiment-compatible contacts, doing so for every demonstration is slow and costly. The challenge is therefore to retain the scalability of human demonstrations while providing the demonstrator with robot-specific contact feedback. 
We ask: \emph{Can humans see the target robot's contact state during demonstration collection and adjust their motions accordingly, without executing the physical robot?}

Existing approaches address this challenge through three main routes.
Human-centric collection scales readily but does not expose the target robot's contact states
\cite{osmo,pmlr-v270-cheng25b}.
Robot teleoperation directly captures interactions with the target embodiment, but requires costly robot hardware for every demonstration
\cite{Qin2023AnyTeleopAG,humanagentjl,xue2026tubediffusionpolicyreactive}.
Human-to-robot retargeting maps human motion to robot configurations and can visualize the retargeted hand during collection
\cite{yin2025geometricretargetingprincipledultrafast,Malate2026SmoothOA,
mandi2026dexmachina,arcap,nechyporenko2024armadaaugmentedrealityrobot}.
However, visualizing the retargeted hand does not reveal whether, where, or when the target hand will establish contact with the object. A motion may appear kinematically plausible while producing contacts that are unstable, missing, or infeasible. The missing ingredient is therefore \emph{target-hand contact feedback integrated into a scalable human demonstration loop}.

\begin{figure}[t]
\centering
\includegraphics[width=0.49\textwidth]{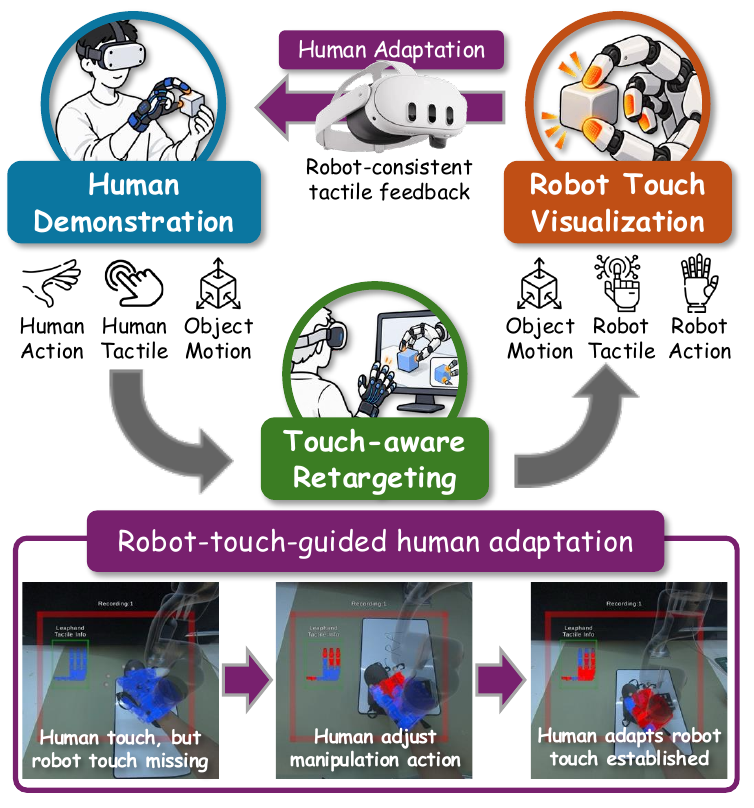}
\caption{\textbf{Touch2Robot overview.}
Human motion, tactile measurements, and object motion condition object-specific RL policies, which are distilled into a unified real-time retargeter.
Retargeted robot configurations and tracked object poses are synchronized in simulation to reconstruct target-hand contacts, enabling online human adaptation.}
\label{fig:teaser}
\vspace{-0.55cm}
\end{figure}

As shown in Fig.~\ref{fig:teaser}, we introduce \textbf{Touch2Robot}, a framework that provides humans with target-hand contact feedback during demonstration collection without requiring physical robot execution. Our key insight is to treat human touch as a \emph{cross-embodiment contact preference}: object motion specifies \emph{what} should happen, whereas human touch guides \emph{how} the robot should interact with the object. Touch2Robot learns object-specific contact-aware RL policies conditioned on human motion, tactile measurements, and object motion, and distills these policies into a unified real-time retargeter. During collection, the retargeted robot hand and tracked object are simulated together to reconstruct the contacts that the target hand would establish. The human then observes these reconstructed contacts and can adapt the ongoing demonstration to avoid unstable, missing, or infeasible contacts.

We evaluate Touch2Robot on four real-world dexterous manipulation tasks. Our experiments examine three aspects of the framework: whether reconstructed target-hand contacts agree with real-robot tactile measurements and human contact preferences, whether robot-touch feedback improves the transferability and collection efficiency of human demonstrations, and whether these gains translate to downstream imitation learning. Across these evaluations, Touch2Robot consistently outperforms visual-only collection and interaction-aware retargeting baselines, demonstrating that target-hand touch provides useful feedback for adapting human demonstrations to the robot embodiment.

Our contributions are:
\begin{itemize}
\item We introduce \textbf{Touch2Robot}, a human demonstration framework that provides target-hand contact feedback for online adaptation without requiring physical robot execution for every demonstration.
\item We develop a contact-aware RL retargeting framework that treats human touch as a cross-embodiment contact preference and object motion as the desired outcome. Object-specific behaviors are distilled into a unified real-time retargeter, enabling target-hand contact reconstruction through synchronized simulation.
\item Across four real-world tasks, we demonstrate accurate contact reconstruction, improved replay success through contact feedback, and higher downstream imitation-learning success.
\end{itemize}

\section{RELATED WORKS}
\subsection{Tactile-Rich Human Data Collection}

Human-centric data collection enables robots to learn manipulation skills from natural human demonstrations. Early systems primarily capture human motion and provide robot-aware guidance: DexCap~\cite{dexcap} records hand motion and scene observations, while ARCap~\cite{arcap} and ARMADA~\cite{nechyporenko2024armadaaugmentedrealityrobot} visualize retargeted robot motions to help demonstrators produce robot-compatible actions. Recent approaches enrich these demonstrations with contact information. DexViTac~\cite{Chen2026DexViTacCH} captures visuo-tactile-kinematic demonstrations, while DexUMI~\cite{Xu2025DexUMIUH} and DEXOP~\cite{fang2025dexopdevicerobotictransfer} provide contact feedback through wearable or mechanically coupled interfaces. RealDexUMI~\cite{Xu2026RealDexUMIAW} further reduces the embodiment gap by sharing a dexterous end-effector between data collection and deployment. Despite these advances, existing systems do not directly provide the demonstrator with the predicted contact state of an independently retargeted robot hand. Touch2Robot addresses this gap by estimating and visualizing target-hand contact online during human demonstration, enabling natural manipulation while providing robot-specific contact guidance.

\subsection{Interaction-Aware Dexterous Retargeting}

Dexterous retargeting transfers human hand movements to a robot hand while accounting for differences in embodiment and kinematics. 
Pose-based methods map human hand configurations to feasible robot poses, typically through fingertip or joint correspondence~\cite{8794277,9197124,10160547}. Interaction-aware methods further preserve hand--object relationships: TopoRetarget~\cite{Wu2026TopoRetargetIR} transfers contact topology, while DexMachina~\cite{mandi2026dexmachina} and ConTrack~\cite{Liang2026ConTrackCH} learn to reproduce object motion through robot contacts. TeleDexter~\cite{li2026teledexter} extends learned interaction control to online teleoperation through consecutive hand--object co-tracking subgoals. ReForce~\cite{Wu2026ReForceLF} learns force-aware residual corrections for online teleoperation and offline demonstration transfer.
These methods improve physical interaction transfer, but do not directly use
the demonstrator's tactile contact pattern as a retargeting preference.
Touch2Robot instead introduces \emph{tactile-aware retargeting}, which incorporates measured human tactile signals as an explicit preference for robot contact realization. It further visualizes the resulting target-hand contacts online, allowing the demonstrator to adapt to the robot's contact behavior during collection.

\section{METHOD}
\begin{figure*}[t]
\centering
\includegraphics[width=\textwidth]{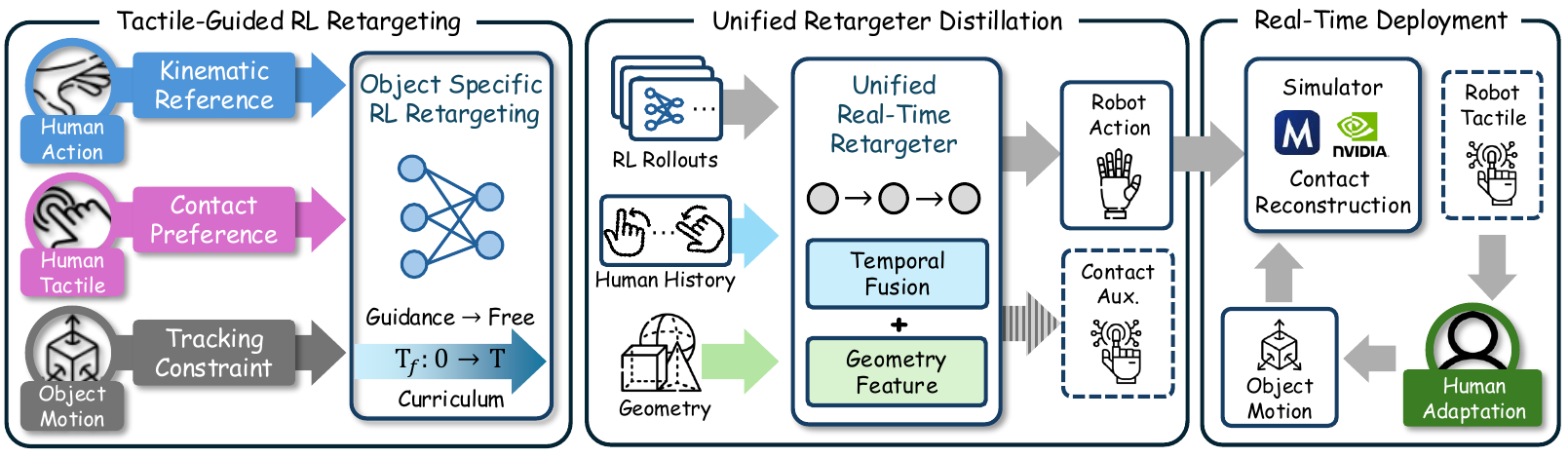}
\caption{\textbf{Touch2Robot pipeline.} Human motion, tactile preference, and
object motion train object-specific RL teachers to learn physically feasible
robot interactions. Their behaviors are distilled into a unified,
geometry-conditioned retargeter. During collection, simulator-based contact
reconstruction provides target-hand feedback for online human adaptation.}
\label{fig:pipeline}
\vspace{-0.275cm}
\end{figure*}

Touch2Robot enables the collection of robot-compatible motion and contact data directly from human demonstrations through three stages (Fig.~\ref{fig:pipeline}). First, it trains object-specific RL teachers to reproduce the demonstrated object motion while preserving the human contact preference (Sec.~\ref{sec:retargeting}). Second, it distills these teachers into a geometry-conditioned retargeter that maps online human observations to robot behavior (Sec.~\ref{sec:distillation}). Third, the retargeter predicts target hand contact during demonstration collection and provides real-time visual cues for online action adjustment, guiding the demonstrator toward more effective data collection (Sec.~\ref{sec:closed_loop}).

\subsection{Tactile-Guided RL Retargeting}
\label{sec:retargeting}

The first stage learns how a demonstrated human interaction can be realized by
the target robot hand. During human manipulation, we observe a trajectory with tactile perception:
\begin{equation}
\mathcal{D}_H
=
\left\{
\left(
\mathbf{q}^{H}_{t},
\boldsymbol{\tau}^{H}_{t},
\mathbf{x}^{o}_{t}
\right)
\right\}_{t=1}^{T},
\end{equation}
where $\mathbf{q}^{H}_{t}$, $\boldsymbol{\tau}^{H}_{t}$, and
$\mathbf{x}^{o}_{t}$ denote human hand motion, tactile observations, and the
object 6D pose, respectively. Our goal is to learn a physically feasible robot
realization that reproduces the demonstrated object motion while preserving
the human contact strategy. 

\noindent \textbf{Stage 1: Tactile-preference-guided interaction learning.} 
Since human and robot hands have different sensing
layouts, we first establish a shared semantic representation for comparing
their contact patterns.
We map both hands to corresponding semantic regions, such as fingertips, palm,
and side-contact regions. For a target hand with $K$ regions, the contact
states of both hands are represented as binary vectors
$\mathbf{c}^{H}_{t},\mathbf{c}^{R}_{t}\in\{0,1\}^{K}$, where each entry
indicates whether the corresponding semantic region is in contact. In this
shared representation, $\mathbf{c}^{H}_{t}$ defines a
\emph{tactile preference}, specifying where the robot should establish and
maintain contact. To represent when contacts appear, persist, or disappear, we encode contact transitions between consecutive
frames:
\begin{equation}
\Delta\mathbf{c}^{H}_{t}
=
\mathbf{c}^{H}_{t}
-
\mathbf{c}^{H}_{t-1},
\qquad
\Delta\mathbf{c}^{R}_{t}
=
\mathbf{c}^{R}_{t}
-
\mathbf{c}^{R}_{t-1}.
\label{eq:contact_transition}
\end{equation}
The tactile reward jointly matches the instantaneous contact pattern and its
temporal evolution:
\begin{equation}
r^{\mathrm{tac}}_t
=
-D\!\left(
\mathbf{c}^{H}_{t},
\mathbf{c}^{R}_{t}
\right)
-
\lambda_{\Delta}
D\!\left(
\Delta\mathbf{c}^{H}_{t},
\Delta\mathbf{c}^{R}_{t}
\right),
\label{eq:tactile_reward}
\end{equation}
where $D({u},{v})=\frac{1}{K}\sum_{i=1}^{K}|u_i-v_i|$
computes the mean absolute difference over the $K$ corresponding regions and $\lambda_{\Delta}$ weights the temporal term. The first term matches the instantaneous contact state, while the second encourages consistent contact onset and release timing throughout the interaction.

\noindent \textbf{Stage 2: Training Object-specific RL teachers.}
Given the tactile preference, we train one RL teacher for each object to exploit the contact modes from each
object geometry. Following the constrained tracking formulation of
ConTrack~\cite{Liang2026ConTrackCH}, we use MTBench~\cite{joshi2025benchmarking}
to optimize multiple trajectories of the same object with a shared teacher.
Specifically, given human motion converted into a robot-side reference
$\mathbf{q}^{\mathrm{ref}}_{1:T}$, and the demonstrated object trajectory
$\mathbf{x}^{o,\mathrm{ref}}_{1:T}$ as the tracking target. The policy produces a residual action relative to the reference:
\begin{equation}
\mathbf{q}^{\mathrm{tar}}_t
=
\mathbf{q}^{\mathrm{ref}}_t+\mathbf{a}^{R}_t,
\end{equation}
where $\mathbf{a}^{R}_t$ is the RL action and $\mathbf{q}^{\mathrm{tar}}_t$ is the resulting robot joint target. 
To allow contact-driven adjustments without excessive deviation from
the reference, we penalize residuals beyond a joint-specific threshold:
\begin{equation}
    r^{\mathrm{act}}_t
    =
    -\sum_i
    \left[
        \max\!\left(0, |a^{R}_{t,i}|-a_i^{\max}\right)
    \right]^2,
    \label{eq:action_penalty}
\end{equation}
where $a_i^{\max}$ defines the penalty-free residual range for joint $i$.
This term permits adjustments within the range rather than penalizing
every deviation from the reference configuration.

To encourage accurate object tracking, we reward agreement between the simulated and demonstrated object motion:
\begin{equation}
    r^{\mathrm{obj}}_t
    =
    -w_p\left(e^{\mathrm{pos}}_t\right)^2
    -w_r\left(e^{\mathrm{rot}}_t\right)^2,
    \label{eq:motion_reward}
\end{equation}
where $e^{\mathrm{pos}}_t$ and $e^{\mathrm{rot}}_t$ denote the Euclidean position error and relative rotation angle between the simulated and reference object poses at time $t$, respectively. The positive weights $w_p$ and $w_r$ balance position and orientation tracking.

The total reward combines object-motion tracking, tactile preference,
and residual-action regularization:
\begin{equation}
    r_t
    =
    \lambda_1 r^{\mathrm{obj}}_t
    +
    \lambda_2 r^{\mathrm{tac}}_t
    +
    \lambda_3 r^{\mathrm{act}}_t,
    \label{eq:style_reward}
\end{equation}
where $\lambda_1,\lambda_2,\lambda_3>0$ balance the three terms.
This objective prioritizes reproducing the demonstrated object motion through human-preferred contacts, while discouraging excessive corrections to the robot-side reference.

\noindent \textbf{Stage 3: Guided-to-free dynamics curriculum.}
In early stage of training, small action errors can drive the object away from the demonstrated trajectory, making subsequent contact learning difficult. We therefore introduce a guided-to-free dynamics curriculum
that gradually replaces reference-state guidance with free robot--object
interaction. During the free-dynamics portion, the robot acts on the object
while its state is advanced purely by simulator physics, without correction
toward the demonstrated trajectory.

For each demonstration, the trajectory is divided into a free-dynamics prefix and a reference-guided suffix:
\begin{equation}
\mathbf{x}^{o}_{t+1} =
\begin{cases}
F_{\mathrm{dyn}}\!\left(
\mathbf{x}^{o}_t, \mathbf{x}^{R}_t, \mathbf{a}^{R}_t
\right),
& t+1 \le T_f,\\[2pt]
\mathbf{x}^{o,\mathrm{ref}}_{t+1},
& t+1 > T_f,
\end{cases}
\label{eq:guided_dynamics}
\end{equation}
where $T_f$ marks the end of the free-dynamics prefix and $F_{\mathrm{dyn}}$ denotes one physics simulation step. The object state includes position, orientation, and velocity. For each demonstration, frames up to $T_f$ evolve under free robot--object dynamics, while the remaining frames are reset to their reference states after each control step. Once the policy reliably controls the current free-dynamics prefix, we increase $T_f$, progressively shortening the reference-guided suffix. This process continues until $T_f=T$, when the entire trajectory is executed without any reference-state reset.

\subsection{Unified Retargeter Distillation}
\label{sec:distillation}

The RL teachers learn physically feasible interaction strategies, but each
teacher is tied to a particular object and relies on a precomputed robot-side
reference. This prevents direct deployment for live demonstration collection.
We therefore distill all object-specific teachers into a single model that
directly maps online human observations to robot behavior across different
objects.

Two inputs enable unified retargeting. First, object geometry is encoded by a
pretrained PointNet~\cite{qi2017pointnetdeeplearningpoint} into a feature
$\mathbf{g}^{o}\in\mathbb{R}^{32}$, allowing the retargeter to condition its
behavior on object-specific contact affordances. Second, the recent history of
human motion, tactile preference, and object motion,
denoted by $\mathbf{h}_t$, provides the temporal context needed for online
prediction. A lightweight causal Transformer fuses these inputs using only past observations. The unified retargeter $f_{\theta}$
predicts both the robot target configuration and its contact state:
\begin{equation}
\left(
\hat{\mathbf{q}}^{\mathrm{tar}}_{t},
\hat{\mathbf{c}}^{R}_{t}
\right)
=
f_{\theta}
\left(
\mathbf{h}_{t},
\mathbf{g}^{o}
\right).
\label{eq:unified_retargeter}
\end{equation}

We train the retargeter by distilling rollouts from all object-specific RL
teachers. The student jointly predicts the robot target configuration
$\mathbf{q}^{\mathrm{tar}}_{t}$ and the contact state
$\mathbf{c}^{R}_{t}$ induced by the teacher rollout:
\begin{equation}
\mathcal{L}_{\mathrm{distill}}
=
\left\|
\hat{\mathbf{q}}^{\mathrm{tar}}_{t}
-
\mathbf{q}^{\mathrm{tar}}_{t}
\right\|_2^2
+
\lambda_c
\operatorname{BCE}
\left(
\hat{\mathbf{c}}^{R}_{t},
\mathbf{c}^{R}_{t}
\right).
\label{eq:distillation_loss}
\end{equation}
The first term transfers the teacher's robot joint targets, while the second
transfers the contact states induced by the learned interaction.

After distillation, a single geometry-conditioned model is used as the final retargeter. This resulting retargeter predicts robot behavior directly from
human interaction history and object geometry, enabling real-time deployment
during demonstration collection.

\subsection{Closed-Loop Demonstration Collection}
\label{sec:closed_loop}

The distilled retargeter provides cues to guide human action via visualizing target-hand
contact. During collection, the system records human motion, human touch, and
object pose. It predicts the robot configuration
$\hat{\mathbf{q}}^{\mathrm{tar}}_{t}$, reconstructs the corresponding robot
hand in simulation, and maps the simulated contacts to the target hand's
semantic tactile regions. The resulting binary tactile state
$\tilde{\mathbf{c}}^{R}_{t}$ is returned to the demonstrator as immediate visualization
feedback.

\textbf{Real-time robot feedback.}
The retargeted robot hand and its reconstructed contact pattern are displayed in VR at 30~Hz with 122~ms end-to-end latency. During manipulation, the demonstrator can observe both the motion the robot hand would execute and the regions where it would make contact. If the reconstructed contact differs from the intended interaction, the demonstrator can adjust the subsequent motion, which is retargeted online. This feedback loop makes target-hand contact errors observable and correctable during data collection.

\begin{figure}[t]
    \centering
    \includegraphics[width=\columnwidth]{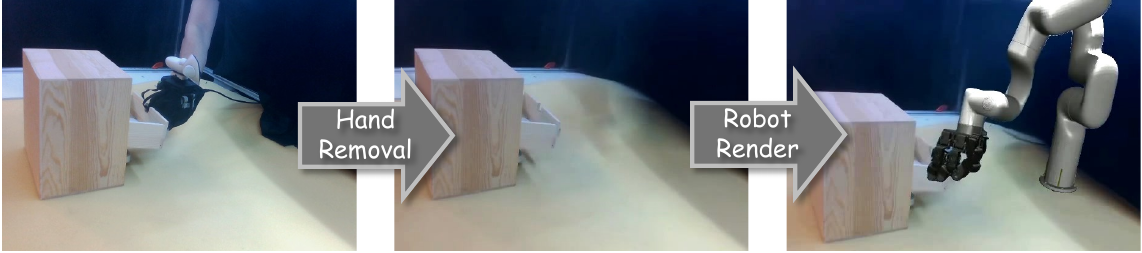}
    \caption{\textbf{Robot-ready visual conversion.}
The human hand is segmented and inpainted, after which the retargeted
robot hand is rendered into the scene using calibrated camera parameters.}
    \label{fig:inpaint}
    \vspace{-10px}
\end{figure}

\textbf{Robot-ready visual observations.}
Downstream visuomotor policies require visual observations consistent with the target robot embodiment. We therefore segment and remove the human hand from RGB frames using SAM~\cite{kirillov2023segment} and ProPainter~\cite{zhou2023propainter}, and render the retargeted robot hand into the inpainted scene using calibrated camera parameters. Fig.~\ref{fig:inpaint} illustrates this visual conversion process. 

\section{EXPERIMENTS}
\begin{figure}[t]
    \centering
    \includegraphics[width=\columnwidth]{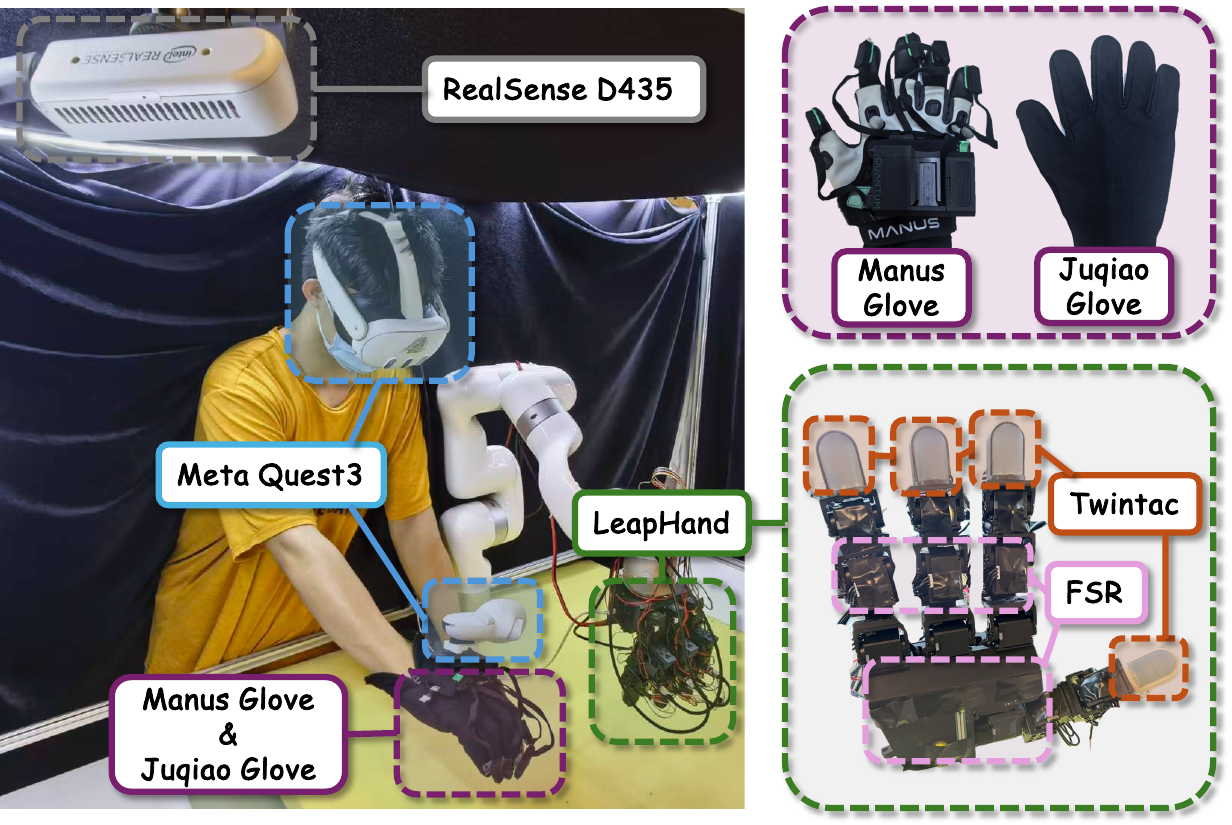}
    \caption{\textbf{Touch2Robot platform.} The system integrates human motion and tactile sensing, object tracking, VR-based robot-touch feedback, and a tactile-enabled dexterous robot for closed-loop demonstration collection and evaluation.}
    \label{fig:platform}
    \vspace{-15px}
\end{figure}

\begin{figure*}[t]
    \centering
    \includegraphics[width=\textwidth]{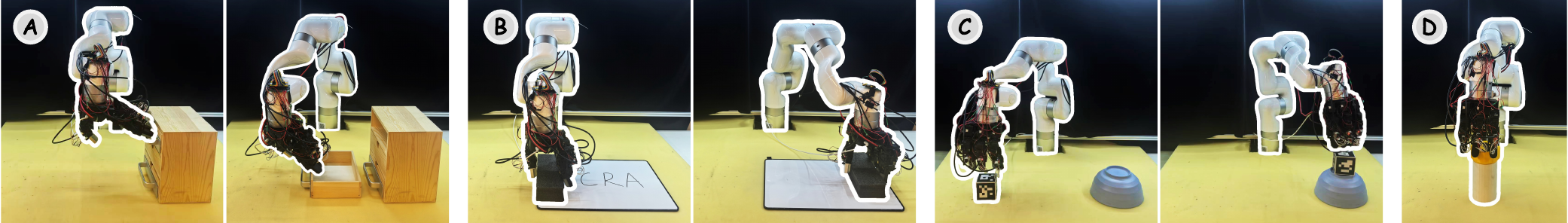}
    \caption{\textbf{Touch2Robot tasks.} We evaluate (A) Pick-and-Place,
    (B) Board Wiping, (C) Object Rotation, and (D) Drawer Opening, covering
    grasp establishment, contact transitions, and sustained-contact interactions.}
    \label{fig:tasks}
    \vspace{-5px}
\end{figure*}

\begin{table*}[t]
    \centering
    \caption{\textbf{Human adaptation and demonstration quality.}
Task columns report real-robot replay completion scores (mean~$\pm$~std, \%). Collection Time denotes the human demonstration time required to obtain one replay-successful trajectory.}
\vspace{-5px}
    \label{tab:demo_quality}

    \setlength{\tabcolsep}{2.5pt}
    \renewcommand{\arraystretch}{1.05}

    \begin{tabularx}{\textwidth}{
    @{}
    >{\columncolor{white}[0.5pt][\tabcolsep]}l
    *{4}{
        >{\hsize=1.0\hsize
          \linewidth=\hsize
          \centering\arraybackslash}X
    }
    >{\hsize=0.8\hsize
      \linewidth=\hsize
      \centering\arraybackslash}X
    >{\columncolor{white}[\tabcolsep][0.5pt]
      \hsize=1.2\hsize
      \linewidth=\hsize
      \centering\arraybackslash}X
    @{}
}
        \toprule
        \textbf{Collection Interface}
        & \textbf{Pick (\%)}
        & \textbf{Rotate (\%)}
        & \textbf{Wipe (\%)}
        & \textbf{Drawer (\%)}
        & \textbf{Average (\%)}
        & \textbf{Collection Time (s)} \\
        \midrule

        Offline Retarget
        & 25.0 $\pm$ 0.0 & 0.0 $\pm$ 0.0 & 15.0 $\pm$ 24.6 & 73.0 $\pm$ 26.3 & 28.3 & 66.0 \\

        Visual Feedback
        & 27.5 $\pm$ 7.9 & 0.0 $\pm$ 0.0 & 49.0 $\pm$ 30.3 & 75.0 $\pm$ 27.6 & 37.9 & 58.6 \\

        \rowcolor{myblue}
        \textbf{Touch2Robot}
        & 62.5 $\pm$ 27.0
        & 59.0 $\pm$ 22.8
        & 90.0 $\pm$ 8.2
        & 77.0 $\pm$ 25.8
        & 72.1
        & 18.2 \\

        \bottomrule
    \end{tabularx}
    \vspace{-5px}
\end{table*}

\begin{table*}[t]
    \centering
    \caption{\textbf{Real-world downstream imitation learning.}
    Diffusion Policy completion scores are reported as mean~$\pm$~std (\%).}
    \vspace{-5px}
    \label{tab:downstream_learning}
    \small
    \setlength{\tabcolsep}{4pt}
    \renewcommand{\arraystretch}{1.05}

    \begin{tabularx}{\textwidth}{
    l
    *{5}{>{\centering\arraybackslash}X}
}
        \toprule
        \textbf{Data Source}
        & \textbf{Pick (\%)}
        & \textbf{Rotate (\%)}
        & \textbf{Wipe (\%)}
        & \textbf{Drawer (\%)}
        & \textbf{Average (\%)} \\
        \midrule

        Offline Retarget
        & 25.0 $\pm$ 0.0
        & 0.0 $\pm$ 0.0
        & 8.0 $\pm$ 4.2 & 60.0 $\pm$ 45.9 & 23.3 \\

        Visual Feedback
        & 27.5 $\pm$ 7.9
        & 0.0 $\pm$ 0.0
        & 39.0 $\pm$ 28.1 & 65.0 $\pm$ 41.2 & 32.9 \\

        \rowcolor{myblue}
        \textbf{Touch2Robot}
        & 52.5 $\pm$ 34.3
        & 43.5 $\pm$ 15.6
        & 83.0 $\pm$ 13.4 & 69.0 $\pm$ 47.7 & 62.0 \\

        \bottomrule
    \end{tabularx}
    \vspace{-10px}
\end{table*}

We evaluate Touch2Robot through three questions:
\textbf{(Q1) Demonstration Quality and Downstream Utility:}
Does online robot tactile feedback improve real-robot replay completion
and downstream imitation-learning performance? (Sec.~\ref{sec:demo_quality})
\textbf{(Q2) Tactile Fidelity and Preference Alignment:}
Do the reconstructed tactile states match real-robot measurements, and do the resulting robot contacts align with human tactile preferences? (Sec.~\ref{sec:tactile_fidelity})
\textbf{(Q3) Feedback Usability:}
Do operators find robot tactile feedback easy to interpret and useful for refining their demonstration collection? (Sec.~\ref{sec:user_study})

\subsection{Experimental Setup}
\label{sec:exp_setup}

\textbf{Hardware and data collection.}
As shown in Fig.~\ref{fig:platform}, our robot platform consists of an xArm6
with a LEAP Hand~\cite{shaw2023leaphand} equipped with TwinTac sensors
~\cite{huang2025twintacwiderangehighlysensitive}. The hand also incorporates
binary FSR sensors attached to the linkages to measure robot tactile signals.
During human demonstration collection, a MANUS glove records hand motion, a
Juqiao tactile glove records human touch, and calibrated RGB cameras track the
object using FoundationPose~\cite{wen2024foundationposeunified6dpose}. A Meta
Quest 3 headset displays the retargeted robot hand and reconstructed contact.

\textbf{Tasks.}
As shown in Fig.~\ref{fig:tasks}, we evaluate four tasks with distinct
contact requirements: \emph{Pick-and-Place}, which requires establishing a
stable grasp and releasing the object within a target region; \emph{Object
Rotation}, requires rotating the object by at least 180$^\circ$ while retaining it in the hand; \emph{Board Wiping}, which requires grasping an
eraser and removing a marked region; and \emph{Drawer Opening}, which requires
grasping the handle and pulling the drawer open. A trial is considered
successful only if the task criterion is satisfied without dropping the
object or triggering a safety stop.


\begin{table*}[t!]
    \centering
    \setlength{\tabcolsep}{3pt}
    \renewcommand{\arraystretch}{1.1}

    \begin{minipage}[t]{0.49\textwidth}
        \centering
        \caption{\textbf{Tactile fidelity.}
        Reconstructed contacts vs. real-robot measurements,
        averaged across four tasks.}
        \label{tab:tactile_fidelity}
        \small
        \vspace{-5px}
         \begin{tabularx}{\linewidth}{
    l*{3}{>{\centering\arraybackslash}X}
}
            \toprule
            \textbf{Method}
            & \shortstack{\textbf{F1}\textbf{(\%)}}
            & \shortstack{\textbf{Onset}\textbf{(ms)}}
            & \shortstack{\textbf{FPR}\textbf{(\%)}} \\
            \midrule
            ConTrack~\cite{Liang2026ConTrackCH}
            & 23.77 & 1383.3 & 4.10 \\
            DexMachina~\cite{mandi2026dexmachina}
            & 24.71 & 1785.9 & 4.74 \\
            Ours w/o Tactile Reward
            & 24.58 & 1694.4 & 2.01 \\
            \rowcolor{myblue}
            \textbf{Ours (Touch2Robot)}
            & 44.19 & 857.4 & 6.38 \\
            \bottomrule
        \end{tabularx}
    \end{minipage}
    \hfill
    \begin{minipage}[t]{0.49\textwidth}
        \centering
        \caption{\textbf{Preference alignment.}
        Reconstructed contacts vs. aligned human tactile preferences,
        averaged across four tasks.}
        \vspace{-5px}
        \label{tab:preference_alignment}
        \small
        \begin{tabularx}{\linewidth}{
    l*{3}{>{\centering\arraybackslash}X}
}
            \toprule
            \textbf{Method}
            & \shortstack{\textbf{F1}\textbf{(\%)}}
            & \shortstack{\textbf{Onset}\textbf{(ms)}}
            & \shortstack{\textbf{FPR}\textbf{(\%)}} \\
            \midrule
            ConTrack~\cite{Liang2026ConTrackCH}
            & 11.38 & 1215.4 & 2.59 \\
            DexMachina~\cite{mandi2026dexmachina}
            & 18.56 & 1453.0 & 3.42 \\
            Ours w/o Tactile Reward
            & 15.26 & 1990.7 & 4.50 \\
            \rowcolor{myblue}
            \textbf{Ours (Touch2Robot)}
            & 47.53 & 291.7 & 6.52\\
            \bottomrule
        \end{tabularx}
    \end{minipage}
    \vspace{-15px}
\end{table*}

\subsection{Demonstration Quality and Downstream Learning}
\label{sec:demo_quality}

\begin{figure}[t]
    \centering
    \includegraphics[width=\columnwidth]{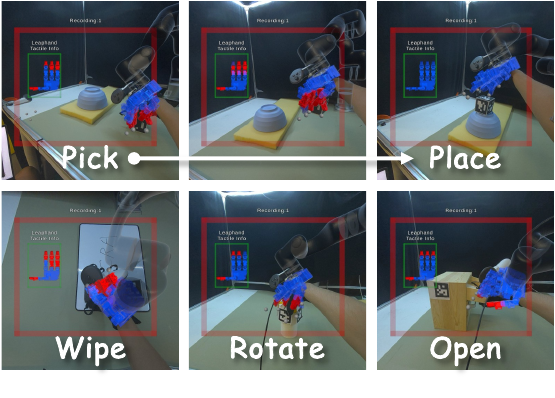}
    \vspace{-20px}
    \caption{\textbf{VR visualization.} Real-time robot motion and tactile
    feedback during human demonstration collection.}
    \label{fig:vr}
    \vspace{-20px}
\end{figure}

We evaluate whether robot tactile feedback improves both the execution quality of collected demonstrations and the performance of policies trained on them (\textbf{Q1}). We compare three collection interfaces using the same unified retargeter and sensing pipeline: \emph{Offline Retargeting}~\cite{Qin2023AnyTeleopAG}, which converts human demonstrations directly to robot trajectories via mapping algorithm; \emph{Visual Feedback}~\cite{arcap}, which displays the retargeted robot hand during collection; and \emph{Touch2Robot}, which additionally displays reconstructed tactile activations. 
Each operator collects 10 attempts for each of the four tasks under all three interfaces. Each collected dataset is used for both replay evaluation and policy training, without filtering or refining demonstrations based on replay outcomes. Fig.~\ref{fig:vr} illustrates the feedback provided during collection.

\textbf{Evaluation metric.}
We use the same task-completion criteria for demonstration replay and downstream policy evaluation. For Object Rotation, Board Wiping, and Drawer Opening, completion is measured as $s=p/p^{\mathrm{goal}}\times100\%$, where $p$ denotes the achieved rotation angle, erased area, or drawer displacement, respectively. For Pick-and-Place, the task is divided into four stages: reach, pick, move, and place, with each completed stage contributing 25\%.

\textbf{Demonstration replay.}
We replay each retargeted robot trajectory on the physical robot from its corresponding initial object configuration and evaluate it using the task completion criteria defined above. Collection time measures the human demonstration time required to obtain one trajectory that successfully completes the task during real-robot replay.

Table~\ref{tab:demo_quality} shows that Touch2Robot substantially improves real-robot replay completion over both Offline Retargeting and Visual Feedback. The largest gains occur on Object Rotation and Board Wiping, where transfer depends strongly on sustained and evolving contacts. Touch2Robot also reduces collection time, indicating more efficient collection of robot-executable demonstrations.

\textbf{Downstream policy learning.}
We next test whether the same improvement transfers to autonomous policy learning. We train Diffusion Policy~\cite{chi2024diffusionpolicyvisuomotorpolicy} using the same number of trajectories per task and data source, and evaluate the learned policies using the same task-completion criteria.

Table~\ref{tab:downstream_learning} reports an average completion score of 62.0\% for Diffusion Policy trained on Touch2Robot demonstrations, compared with 32.9\% for Visual Feedback and 23.3\% for Offline Retargeting.
Together with the replay results, this consistent improvement shows that robot-touch feedback benefits not only the immediate executability of the retargeted demonstrations, but also the supervision they provide for policy learning. These results indicate that exposing target-hand contact during collection helps humans produce more transferable and robot-compatible demonstrations.

\begin{figure}[t]
    \centering
    \includegraphics[width=\columnwidth]{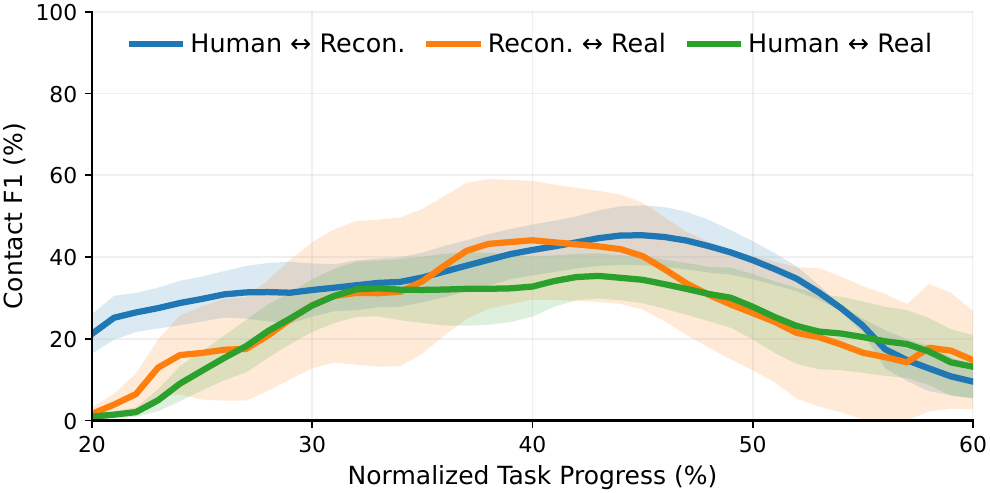}
    \caption{\textbf{Temporal contact agreement.}
Pairwise contact F1 among human tactile preference, reconstructed target-hand
contact, and real-robot tactile measurement.}
    \label{fig:tactile}
    \vspace{-20px}
\end{figure}

\subsection{Tactile Fidelity and Preference Alignment}
\label{sec:tactile_fidelity}

We next investigate \textbf{Q2}: whether the reconstructed target-hand contacts accurately reflect the robot's physical contacts and remain aligned with the human's tactile preference.

\noindent\textbf{Tactile fidelity: does the reconstruction predict real contact?}
We compare the reconstructed target-hand contact sequence with tactile measurements collected when the corresponding retargeted trajectory is replayed on the physical robot. This evaluates whether simulated contact reconstruction accurately predicts the robot's actual contact behavior.

\noindent\textbf{Preference alignment: does retargeting preserve human contact preference?}
We compare the reconstructed target-hand contacts with the semantically aligned human tactile sequence. This evaluates whether tactile-guided retargeting preserves the demonstrated contact pattern and timing.

Tables~\ref{tab:tactile_fidelity} and \ref{tab:preference_alignment} report the two evaluations using contact F1, contact-onset error, and false-positive rate (FPR). In both evaluations, we compare Touch2Robot with ConTrack~\cite{Liang2026ConTrackCH} and DexMachina~\cite{mandi2026dexmachina}, and additionally remove the tactile reward while keeping all other training settings unchanged. These comparisons evaluate the contribution of explicit human tactile supervision beyond interaction-aware retargeting based on motion and geometry.

For tactile fidelity, Touch2Robot achieves 44.19\% F1 while reducing contact-onset error, substantially outperforming both interaction-aware baselines and the variant without tactile reward. The improvement comes with a higher FPR, indicating a trade-off between recovering more true contacts and introducing additional spurious activations.
For preference alignment, Touch2Robot substantially improves agreement with human tactile preferences, achieving 47.53\% F1 while reducing contact-onset error to below 0.3~s. The clear improvement over both interaction-aware baselines and the variant without tactile reward shows the importance of explicit human tactile supervision for realizing the demonstrated contact pattern. The higher FPR suggests that the robot occasionally introduces additional embodiment-specific contacts.
Together, the two evaluations show that tactile supervision improves both contact reconstruction and preference realization, with better contact coverage and timing at the cost of a modest increase in false-positive contacts.

We further examine temporal contact agreement during the main interaction phase of Pick-and-Place. We focus on the 20--60\% normalized task progress, which mainly covers grasp and object transport, and compute pairwise contact F1 among human tactile preference, reconstructed target-hand contact, and real-robot tactile measurements.
As shown in Fig.~\ref{fig:tactile}, the reconstructed contact exhibits comparable agreement with both the human tactile preference and the real-robot measurement throughout most of the interaction. This indicates that Touch2Robot does not improve alignment to one side by sacrificing the other; instead, the reconstructed target-hand contact balances human contact intent with the contact behavior that is physically realized by the robot.

\subsection{User Study on Robot-Touch Feedback}
\label{sec:user_study}

To answer \textbf{Q3}, we conduct a within-subject study to evaluate whether
robot tactile feedback helps people collect better demonstrations. We recruit
10 participants (7 male and 3 female; mean age 24 years), including five with
prior experience in robot teleoperation or dexterous manipulation and five
without. Each participant completes the Rotation task with both
\emph{Visual Feedback} and \emph{Touch2Robot}. After a short familiarization
session, interface order is counterbalanced: half of the participants use
Visual Feedback first, and the other half use Touch2Robot first. Both
conditions use matched initial states and equal demonstration-collection time.

After each interface, participants rate contact awareness, adaptation confidence, feedback usefulness, and ease of use on seven-point Likert scales, where 1 indicates ``strongly disagree'' and 7 indicates ``strongly agree.''
Likert ratings are linearly rescaled to percentages (1=0\%, 7=100\%).
Table~\ref{tab:user_study} shows that Touch2Robot receives substantially higher ratings across all four subjective measures. Participants report better awareness of the robot's contact state, greater confidence in adjusting their motions, and stronger perceived usefulness of the feedback. Ease-of-use ratings also improve. All differences remain significant after Holm correction ($p<0.01$), indicating that robot-touch feedback helps better understand the robot's contact state and improve their motions during collection.

\begin{table}[t]
    \centering
    \caption{\textbf{User study on robot-touch feedback.}
    We evaluate whether robot tactile feedback makes demonstration collection
    more intuitive, useful, and easier to adapt to. }
    \vspace{-5px}
    \label{tab:user_study}
    \small
    \setlength{\tabcolsep}{4pt}
    \renewcommand{\arraystretch}{1.08}

    \begin{tabularx}{\columnwidth}{
        @{}
        >{\raggedright\arraybackslash}p{0.40\columnwidth}
        >{\centering\arraybackslash}X
        >{\centering\arraybackslash}X
        @{}
    }
        \toprule
        \textbf{Metric}
        & \textbf{Visual Feedback}
        & \textbf{Touch2Robot} \\
        \midrule

       Contact Awareness $\uparrow$
& 18.3 $\pm$ 18.3 & 86.7 $\pm$ 13.1 \\

Adaptation Confidence $\uparrow$
& 16.7 $\pm$ 15.7 & 88.3 $\pm$ 11.2 \\

Feedback Usefulness $\uparrow$
& 25.0 $\pm$ 30.7 & 88.3 $\pm$ 8.1 \\

Ease of Use $\uparrow$
& 40.0 $\pm$ 27.4 & 90.0 $\pm$ 8.6 \\

        \bottomrule
    \end{tabularx}
    \vspace{-10px}
\end{table}

\begin{figure}[t]
    \centering
    \includegraphics[width=\columnwidth]{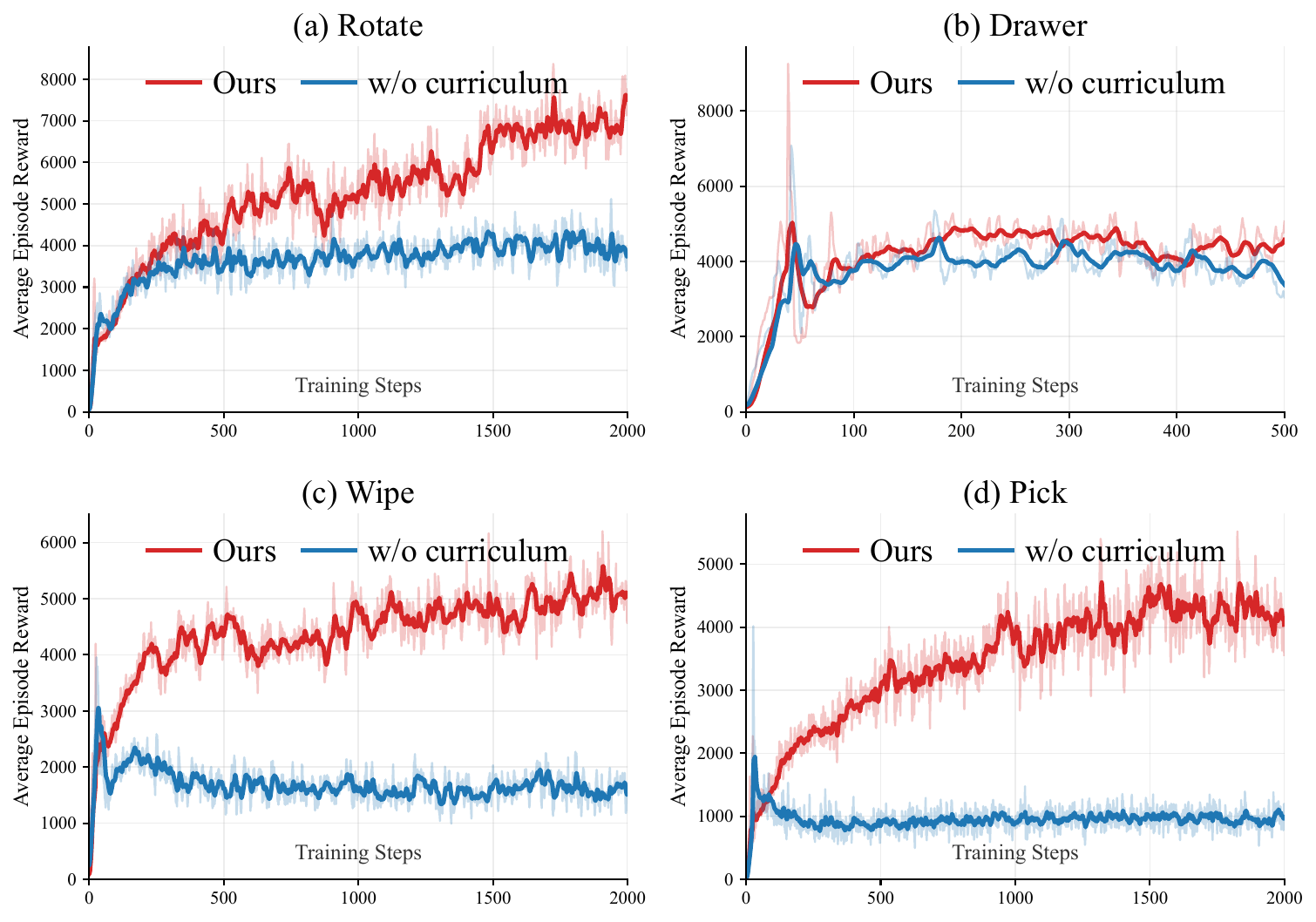}
    \vspace{-10px}
    \caption{\textbf{Effect of the dynamics curriculum.}
Average episode reward during training with and without the curriculum.}
\label{fig:curriculum}
\vspace{-15px}
\end{figure}

\subsection{Algorithm Ablations}
\label{sec:algorithm_ablation}

We study two parts of our learning pipeline. First, we test whether the
dynamics curriculum improves RL teacher training. Second, we test whether
the unified retargeter can match the interaction performance of the
object-specific RL teachers while using a single model. For all interaction
evaluations, the object moves freely according to the simulator dynamics,
without reference-state guidance.

\textbf{Dynamics curriculum.}
We first test whether the dynamics curriculum improves teacher training. We train two teacher variants, with and without the curriculum, and evaluate them at the same training checkpoints. At each checkpoint, both teachers are tested on the same set of complete trajectories. Fig.~\ref{fig:curriculum} reports the average episode reward as a function of simulator training steps. The curriculum consistently achieves higher reward throughout training, indicating more stable and effective teacher optimization. By progressively extending the free-dynamics portion, it reduces the impact of early tracking failures and allows the policy to learn increasingly long-horizon object control before operating fully without guidance.

\textbf{Unified retargeter distillation.}
We next evaluate whether the unified retargeter preserves the interaction quality of the object-specific RL teachers while enabling a single real-time model across objects. We compare it with the corresponding RL teachers under identical simulation conditions, and ablate geometry conditioning and auxiliary contact supervision. Object-motion error is normalized by the reference motion magnitude and reported as a percentage, while contact F1 is computed against human tactile preferences. We also report the per-frame inference time of the unified retargeter.

\begin{table}[t]
    \centering
    \caption{\textbf{Unified retargeter ablation.}
Comparison with object-specific RL teachers and model ablations.}
\vspace{-5px}
    \label{tab:retargeter_ablation}
    \small
    \setlength{\tabcolsep}{3pt}

    \begin{tabularx}{\columnwidth}{
        l*{3}{>{\centering\arraybackslash}X}
    }
        \toprule
        \textbf{Method}
        & \mbox{\textbf{Obj. Err.(\%)}}
        & \textbf{F1(\%) }
        & \textbf{Latency(ms)} \\
        \midrule

        Object-Specific RL 
        & 8.64 & 73.70 & -- \\
        \midrule

        Ours w/o Geometry
        & 24.43 & 51.86 & 1.14  \\

        Ours w/o Contact Aux.
        & 24.03 & 52.73 & 1.12  \\

        \rowcolor{myblue}
        \textbf{Unified Retargeter}
        & 18.05 & 65.46 & 1.12  \\

        \bottomrule
    \end{tabularx}
    \vspace{-10px}
\end{table}

\begin{figure}[t]
    \centering
    \includegraphics[width=\columnwidth]{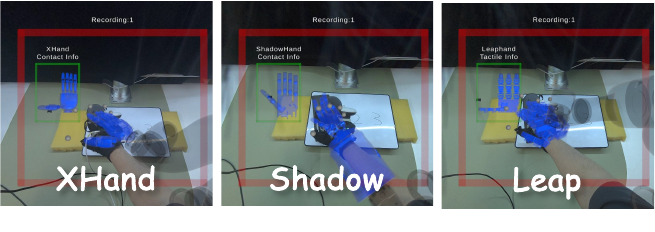}
    \vspace{-20px}
    \caption{\textbf{Cross-hand applicability.}
The same human manipulation is retargeted to three dexterous hand in VR.}
\label{fig:cross_hand}
\vspace{-20px}
\end{figure}

Table~\ref{tab:retargeter_ablation} shows that the unified retargeter preserves most of the interaction quality of the object-specific RL teachers while supporting real-time inference. Removing either geometry conditioning or auxiliary contact supervision degrades both object tracking and contact alignment, indicating that the two components provide complementary cues for distillation. The low inference latency further supports its use for closed-loop demonstration collection.

Together, these ablations show that the curriculum improves teacher learning, while geometry and contact supervision preserve interaction quality during distillation, jointly supporting reliable real-time retargeting.

\subsection{Cross-Hand Applicability}
\label{sec:cross_hand}

Finally, we qualitatively examine whether the Touch2Robot interface can accommodate robot hands with different kinematics and contact layouts. Figure~\ref{fig:cross_hand} visualizes the same human manipulation retargeted to three dexterous hand embodiments in VR. 
This qualitative result illustrates that the Touch2Robot collection loop is not tied to the LEAP Hand interface and can represent robot feedback for different target hands. Quantitative cross-hand evaluation and real-world deployment on additional embodiments remain future work.

\section{CONCLUSION }
We presented \textbf{Touch2Robot}, which brings target-hand contact feedback into scalable human demonstration collection without per-demonstration robot execution. By combining tactile-guided retargeting with online robot-touch feedback, Touch2Robot improves average real-robot replay completion from 37.9\% with visual feedback to 72.1\%, while also improving downstream policy learning. 
A within-subject user study further shows that robot-touch feedback helps users better understand the target hand's contact state and adjust their motions during collection. Qualitative cross-hand results suggest that the same framework can accommodate robot hands with different kinematics and tactile layouts.
Future work will quantitatively validate this capability across additional embodiments and extend the current binary contact representation to richer tactile signals. More broadly, this framework provides a scalable bridge from human manipulation data to robot-ready motion and contact supervision, supporting robot learning from large-scale human demonstrations.

\section*{ACKNOWLEDGMENT}
This work was supported by the National Natural Science Foundation of China (Grant No. 52305007), the Natural Science Foundation of Shanghai (Grant No. 25ZR1402370), the Artificial Intelligence Project of the State Key Laboratory of General Artificial Intelligence, BIGAI, Peking University, Beijing, China (Project No. SKLAGI2025OP19), the State Key Laboratory of Mechanical System and Vibration (Grant No. MSV202519) and the MoE Key Laboratory of Intelligent Perception and Human-Machine Collaboration (KLIP-HuMaCo).

\bibliographystyle{IEEEtranN}
\bibliography{references}

\clearpage

\twocolumn[
\begin{center}
    {\LARGE\bfseries Supplementary Material of Touch2Robot}
\end{center}
\vspace{2.5em}
]

Here we lay down the details of the data collection, training, and testing process. More technical details are given here to illustrate our method and implementations better.

\setcounter{section}{0}
\renewcommand{\thesection}{\Alph{section}}

\section{Additional Method Details}
\label{app:method}
This section provides details on tactile processing, task definitions and rewards, and model training.

\subsection{Tactile Processing and Semantic Alignment}

Touch2Robot represents human, simulated-robot, and real-robot tactile observations in a shared binary semantic contact space. We first process each sensing modality independently and then align corresponding human and robot contact regions for retargeting and evaluation.

\begin{figure}[t]
    \centering
    \begin{minipage}[t]{0.48\linewidth}
        \centering
        \vspace{0pt}
        \includegraphics[
            height=3.6cm,
            width=\linewidth,
            keepaspectratio
        ]{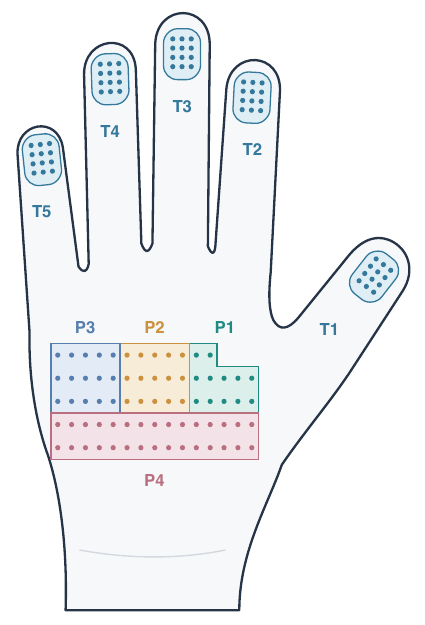}
        \caption{Human tactile-region.}
        \label{fig:tactile_layout}
    \end{minipage}
    \hfill
    \begin{minipage}[t]{0.48\linewidth}
        \centering
        \vspace{0pt}
        \includegraphics[
            height=3.6cm,
            width=\linewidth,
            keepaspectratio
        ]{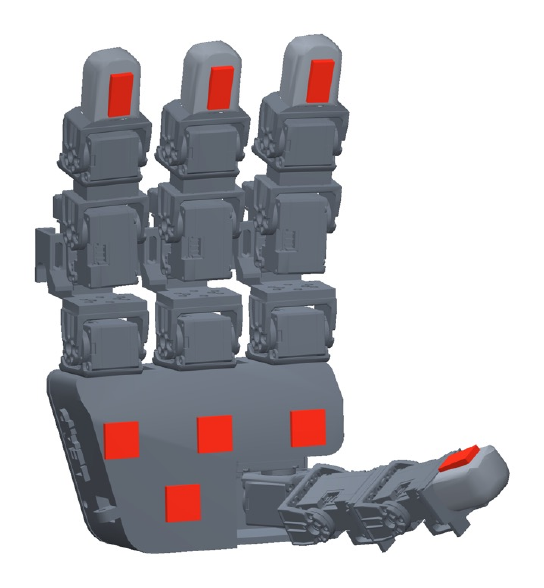}
        \caption{Leap tactile-region.}
        \label{fig:leapfsr}
    \end{minipage}
\end{figure}

The human glove provides 256 taxel readings, which are grouped into semantic fingertip and palm regions according to the glove layout in Fig.~\ref{fig:tactile_layout}. Regional tactile responses are binarized using recording-specific thresholds, with separate thresholds for fingertip and palm regions to account for their different signal ranges. For cross-embodiment alignment with LEAP, we use four fingertip regions corresponding to the thumb, index, middle, and ring fingers, together with four palm regions. The remaining human fingertip region has no robot counterpart and is excluded from contact comparison.

For simulated robot contact, we define eight semantic contact regions corresponding to the LEAP tactile layout: four fingertip regions and four palm regions, as shown in Fig.~\ref{fig:leapfsr}. Each robot region is paired with its corresponding human semantic region. The thumb, index, middle, and ring fingertip sensors correspond to human regions $T1$--$T4$, respectively. The four palm sensors correspond to human regions $P1$--$P4$. We set the simulated force threshold to 0.5 N for both fingertip and palm regions.

For real robot contacts, we record the raw tactile readings together
with their timestamps. We set a channel specific threshold to handle varying baselines and noise levels.
For the fingertip TwinTac channels, thresholds are obtained from an unloaded
calibration recording. The FSR stream uses a zero threshold in its native output units.

\subsection{Task Environments and PPO Training}
\begin{figure}[t]
    \centering
    \includegraphics[width=0.45\textwidth]{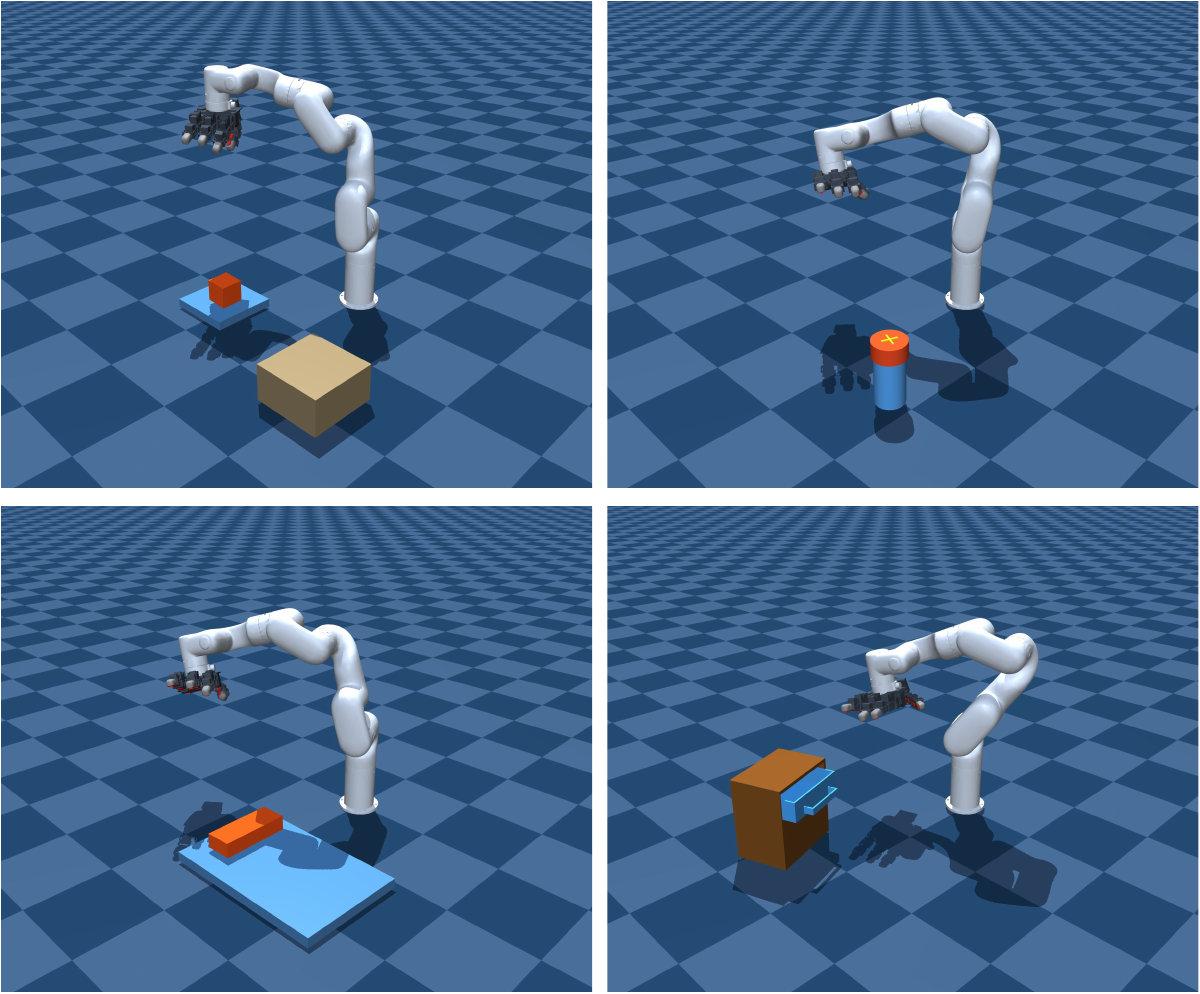}
    \caption{Simulated task settings.}
    \label{fig:simtask}
    \vspace{-10px}
\end{figure}

We evaluate Touch2Robot on four contact-rich manipulation tasks \textit{Pick}, \textit{Rotate}, \textit{Wipe} and \textit{Drawer}. Fig.~\ref{fig:simtask} shows the four LEAP Hand environments in Mujoco. The PPO training hyperparameters are summarized in Table~\ref{tab:ppo_hyperparameters}.

\begin{table}[h]
    \centering
    \caption{PPO training hyperparameters.}
    \label{tab:ppo_hyperparameters}
    \small
    \setlength{\tabcolsep}{5pt}
    \renewcommand{\arraystretch}{1.08}
    \begin{tabular}{lc}
        \toprule
        \textbf{Hyperparameter} & \textbf{Value} \\
        \midrule
        Parallel environments & 200 \\
        Rollout length & 512 \\
        Number of mini-batches & 4 \\
        Optimization epochs & 5 \\
        Hidden dimensions & \([1024,1024,512]\) \\
        Activation & ELU \\
        Learning rate & \(3\times10^{-4}\) \\
        PPO clip range & \(0.2\) \\
        Maximum gradient norm & \(1.0\) \\
        Discount factor \(\gamma\) & \(0.96\) \\
        GAE parameter \(\lambda_{\mathrm{GAE}}\) & \(0.95\) \\
        Initial action-noise std. & \(0.8\) \\
        Desired KL divergence & \(0.016\) \\
        Entropy coefficient & \(0\) \\
        \bottomrule
    \end{tabular}
\end{table}

\begin{figure*}[t]
    \centering
    \includegraphics[width=\textwidth]{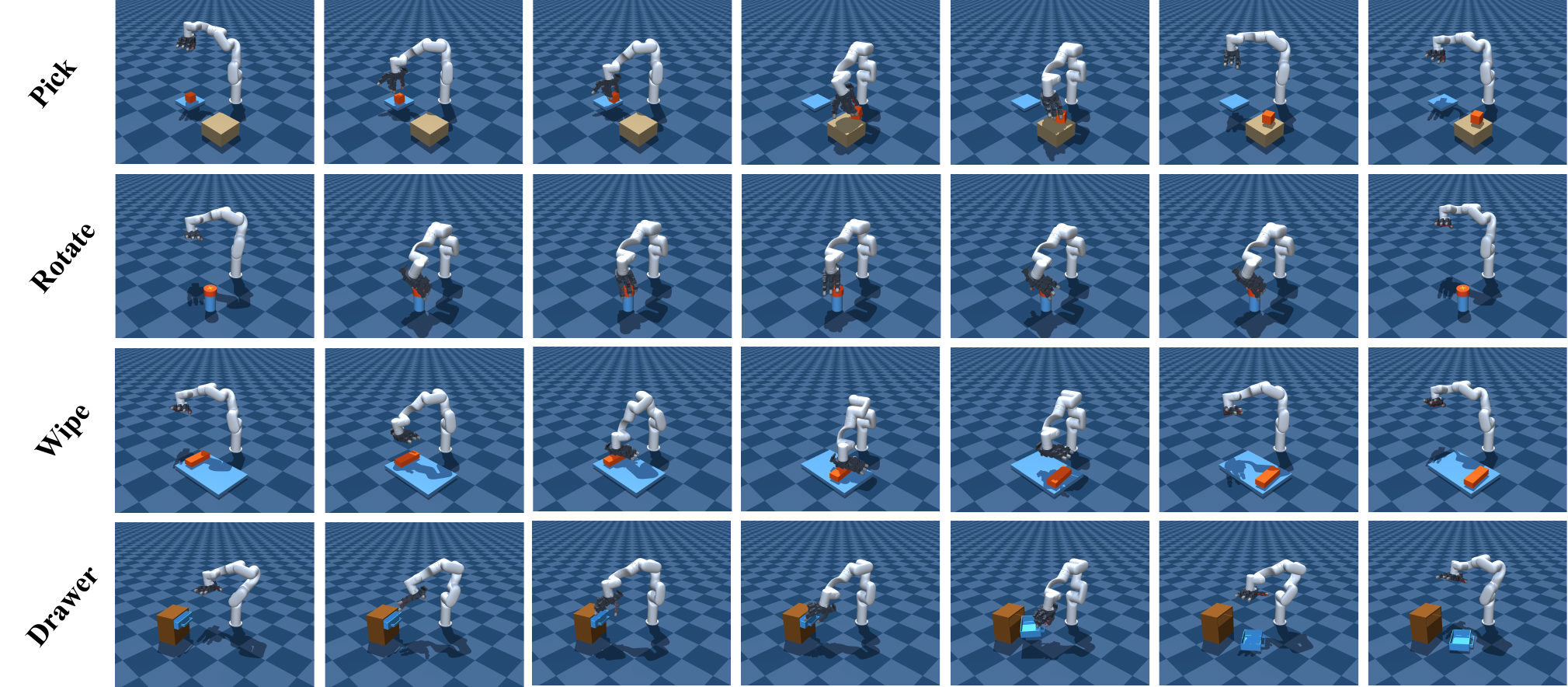}
    \caption{\textbf{Task execution sequences.}
Example rollouts for Pick-and-Place, Rotate, Wipe, and Drawer.}
    \label{fig:tasks}
\end{figure*}

\subsection{Retargeter Distillation}
We construct paired human robot sequences from teacher trajectories, with 30 trajectories for each task. We add Gaussian noise $\epsilon \sim \mathcal{N}(0,0.1^2)$ to the joint positions. The same joint noise are applied to both the human input and the robot teacher target, preserving their frame-wise correspondence, while the object geometry and scene configuration remain unchanged. We collect 30 teacher trajectories per task and augment them to a total of 400 training rollouts.

At each timestep, the student uses a causal Transformer to fuse a recent history of raw human motion retargeting, object poses, tactile preferences. The geometry feature is produced by a PointNet style encoder. The student is supervised using the teacher's robot joint targets and binary contact labels. The architecture and training settings are summarized in Table~\ref{tab:retargeter_hyperparameters}.

\begin{table}[t]
    \centering
    \small
    \caption{Unified retargeter training hyperparameters.}
    \setlength{\tabcolsep}{5pt}
    \renewcommand{\arraystretch}{1.05}
    \begin{tabular}{@{}lc@{}}
        \toprule
        \textbf{Hyperparameter} & \textbf{Value} \\
        \midrule
        History length $H$ & $16$  \\
        Geometry feature dimension &  $32$ \\
        Transformer layers / heads & $4$ / $8$ \\
        Tactile input dropout & $0.25$ \\
        Optimizer & AdamW \\
        Batch size & $128$ \\
        Learning rate & $3{\times}10^{-4}$ \\
        \bottomrule
    \end{tabular}
    \label{tab:retargeter_hyperparameters}
\end{table}

\subsection{Diffusion Policy Training}

For each task, we train downstream policies on robot demonstrations
obtained using each compared data-collection method. Each demonstration contains synchronized RGB observations from two camera views, robot joint states, tactile readings, and joint-target commands.
We use the LeRobot implementation of Diffusion Policy and optimize the standard noise-prediction objective. At deployment, the policy predicts a sequence of joint targets and executes the first 16 actions before replanning from the latest observations. The main training settings are summarized in Table~\ref{tab:dp_training}; all remaining architecture and diffusion settings follow the default configuration.
\begin{table}[h]
    \centering
    \small
    \caption{Diffusion Policy settings.}
    \setlength{\tabcolsep}{4pt}
    \renewcommand{\arraystretch}{0.95}
    \setlength{\aboverulesep}{1.5pt}
    \setlength{\belowrulesep}{1.5pt}
    \begin{tabular}{@{}cc@{}}
        \toprule
        \textbf{Parameter} & \textbf{Value} \\
        \midrule
        Sampling frequency & 20\,Hz \\
        RGB views / resolution & 2 / $224\times224$ \\
        Observation steps & 5 \\
        Prediction horizon & 24 \\
        Exec. steps & 16 \\
        \midrule
        Optimizer & Adam \\
        Learning rate & $10^{-4}$ \\
        Weight decay & $10^{-6}$ \\
        Batch size & 64 \\
        Training updates & 50,000 \\
        \bottomrule
    \end{tabular}
    \label{tab:dp_training}
\end{table}

\section{Baseline Implementation Details}
\label{app:baselines}

This section describes how the retargeting baselines are adapted to our
human demonstration data and LEAP-Hand evaluation setting. Unless otherwise
specified, all methods use the same human motion, object trajectories,
robot model, and simulation environment. Human tactile measurements are
provided only to Touch2Robot.

\begin{table*}[t]
    \centering
    \caption{\textbf{User-study questionnaire.}}
    \label{tab:user_questionnaire}
    \small
    \setlength{\tabcolsep}{3pt}
    \renewcommand{\arraystretch}{1.0}
    \begin{tabularx}{\textwidth}{@{}lp{0.20\textwidth}X@{}}
        \toprule
        \textbf{ID} & \textbf{Dimension} & \textbf{Question} \\
        \midrule
        Q1 & Contact Awareness
        & I could identify which parts of the robot hand
          were in contact with the object. \\

        Q2 & Adaptation Confidence
        & I felt confident in adjusting my movements when the robot hand's
  contact was not as intended. \\

        Q3 & Feedback Usefulness
        & The interface feedback helped me judge whether the demonstration was readily for execution by the robot. \\

        Q4 & Ease of Use
        & I could use the interface feedback without noticeably
          interfering with my natural movements. \\
        \bottomrule
    \end{tabularx}
\end{table*}

\subsection{Retargeting Baselines}

\textbf{Dex Retargeting.}
We use Dex Retargeting as a purely kinematic baseline.
The MANUS hand motion is first transformed into the robot reference frame,
after which the human wrist pose and fingertip geometry are mapped to the
LEAP Hand through frame-wise constrained optimization. Robot joint limits
are enforced during optimization, and the solution from the previous frame
is used to initialize the next frame for temporal continuity.
The resulting LEAP joint trajectory is used directly as the retargeted
robot motion. This baseline does not use object dynamics or human tactile
measurements.

\textbf{ConTrack.}
ConTrack requires robot-side hand-motion, object-motion, and geometric
contact references. We therefore preprocess each human demonstration into
the corresponding LEAP-Hand representation. Human hand motion and object
poses are first transformed into a common metric coordinate frame and
temporally aligned. We then solve sequential, joint-limit-constrained IK
to obtain a LEAP joint reference from the human wrist pose and fingertip
positions.

To construct the contact-style reference required by ConTrack, we extract
human--object proximity from the reconstructed MANO hand. We divide the
fingers into 15 segments and mark a segment as contacting when any of its
vertices lies inside the object or within 5\,mm of its surface. The closest
object-surface point is recorded in the object-local frame and associated
with the corresponding LEAP finger link. The resulting LEAP joint
trajectory, demonstrated object trajectory, and geometry-derived contact
references are then used to train ConTrack in our simulator.

We retain ConTrack's original adaptive task--style optimization, but replace
its original demonstration source with the above references derived from
our human data. Importantly, ConTrack does not receive the measured tactile
glove signals used by Touch2Robot; its contact supervision is obtained only
from hand--object geometry.

\textbf{DexMachina.}
DexMachina is adapted from its original functional-retargeting formulation
to our single-hand LEAP setting. We use the same demonstrated object
trajectory as the task reference and construct the robot-side kinematic
reference from the human motion using the same coordinate alignment and IK
procedure described above. Demonstration-derived contact targets are
obtained geometrically from the reconstructed human hand and object, rather
than from the tactile glove.

We retain DexMachina's task-tracking, imitation, and contact objectives,
together with its virtual-object-assistance curriculum. Terms associated
with the second hand in the original bimanual formulation are removed for
our single-hand tasks. The virtual assistance is progressively reduced
during training until the object evolves entirely under robot--object
dynamics.

\section{Additional Evaluation Details}
\label{app:evaluation}
This section provides additional details on the evaluation procedures used in the main paper. We first define the contact-based metrics for tactile fidelity and preference alignment, and then describe the contact-onset metric and user-study protocol.

\subsection{Contact Precision, Recall, F1, and False-Positive Rate}
We compare binary contact sequences defined over the shared semantic regions. Let $y_{t,c}\in\{0,1\}$ denote the reference contact state and $\hat{y}_{t,c}\in\{0,1\}$ the evaluated contact state, where $t$ is the time index and $c$ is the semantic region.
For tactile fidelity, $y_{t,c}$ is the measured real-robot contact and $\hat{y}_{t,c}$ is the reconstructed target-hand contact.
For preference alignment, $y_{t,c}$ is the aligned human tactile preference and $\hat{y}_{t,c}$ is the reconstructed target-hand contact.

We count true positives ($TP$) when both sequences indicate contact, false positives ($FP$) when only the evaluated sequence indicates contact, false negatives ($FN$) when only the reference indicates contact, and true negatives ($TN$) when neither indicates contact.
We aggregate these counts over all time--region pairs before computing the metrics.

Precision and recall are
\begin{equation}
\mathrm{Precision}=\frac{TP}{TP+FP},
\qquad
\mathrm{Recall}=\frac{TP}{TP+FN}.
\end{equation}

We compute F1 and the false-positive rate as
\begin{equation}
\mathrm{F1}=\frac{2TP}{2TP+FP+FN},
\qquad
\mathrm{FPR}=\frac{FP}{FP+TN}.
\end{equation}

\subsection{Contact-Onset Error}

We align contact sequences using the recorded replay timestamps and their mapping to reference-trajectory time.  For each trajectory and semantic region, we compare the contact-onset times.
Let $t_i^{\mathrm{ref}}$ and $t_i^{\mathrm{eval}}$ denote the corresponding
onset times in seconds. A trajectory--region pair is considered valid when
both sequences contain a contact onset. For $N$ valid pairs, the mean
contact-onset error is
\begin{equation}
E_{\mathrm{onset}}
=
\frac{1000}{N}
\sum_{i=1}^{N}
\left|t_i^{\mathrm{eval}}-t_i^{\mathrm{ref}}\right|.
\end{equation}
The factor $1000$ converts seconds to milliseconds. Missing contacts are
captured by the contact-overlap metrics and are excluded from the onset
error.

\subsection{User Study Protocol and Questionnaire}

We conduct a within-subject comparison between Visual Feedback and Touch2Robot. Visual Feedback displays the retargeted robot hand without contact information, whereas Touch2Robot additionally visualizes reconstructed target-hand contacts. Participants perform the Rotation task under both conditions following a short familiarization session. Condition order is counterbalanced across participants, while initial states and demonstration-collection time are matched. Participants complete the questionnaire in Table~\ref{tab:user_questionnaire} immediately after each condition.

Ten participants rate each item on a seven-point Likert scale. Statistical significance is evaluated on the original ratings using paired two-sided Wilcoxon signed-rank tests with Holm correction.

\end{document}